\documentclass[11pt]{article}
\usepackage[a4paper,margin=1in]{geometry}

\usepackage{graphicx}
\usepackage{amsmath}
\usepackage{amssymb}
\usepackage{multirow}
\usepackage{rotating}

\usepackage[breaklinks=true,colorlinks]{hyperref}

\usepackage{authblk}
\author[1,2]{Vasileios Sevetlidis\thanks{vasiseve@athenarc.gr}}
\author[2]{George Pavlidis}
\author[1]{Spyridon Mouroutsos}
\author[1]{Antonios Gasteratos}
\affil[1]{Democritus University of Thrace, Xanthi, Greece}
\affil[2]{Athena Research Centre, Xanthi, Greece}

\begin{document}
\date{}

\title{Dens-PU: PU Learning with Density-Based Positive Labeled Augmentation}


\maketitle

\begin{abstract}
    This study proposes a novel approach for solving the PU learning problem based on an anomaly-detection strategy. Latent encodings extracted from positive-labeled data are linearly combined to acquire new samples. These new samples are used as embeddings to increase the density of positive-labeled data and, thus, define a boundary that approximates the positive class. The further a sample is from the boundary the more it is considered as a negative sample. Once a set of negative samples is obtained, the PU learning problem reduces to binary classification. The approach, named Dens-PU due to its reliance on the density of positive-labeled data, was evaluated using benchmark image datasets, and state-of-the-art results were attained.    
\end{abstract}

\section{Introduction}

\begin{figure}[t!]
\centering
\includegraphics[height=6cm]{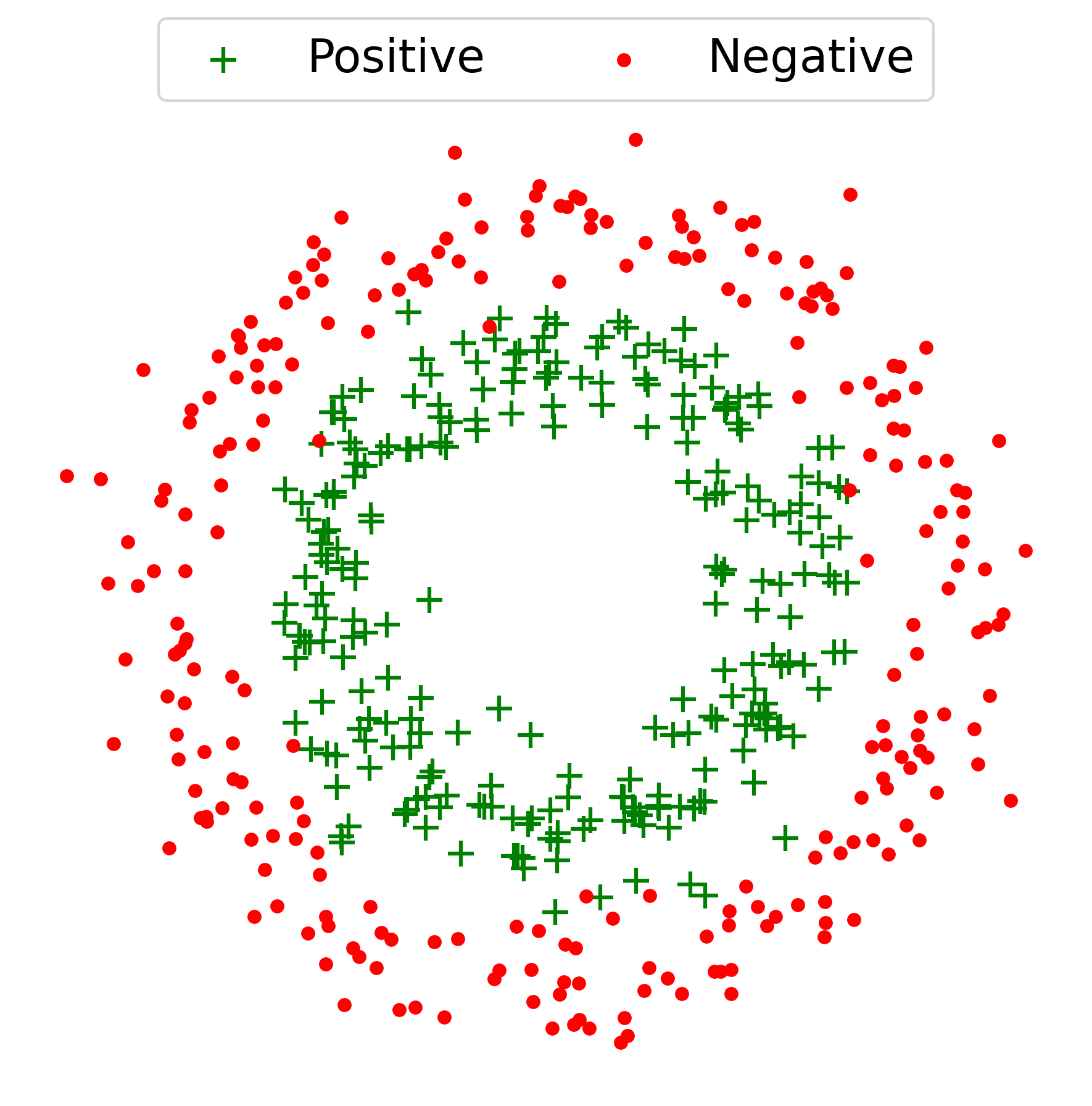}
\includegraphics[height=6cm]{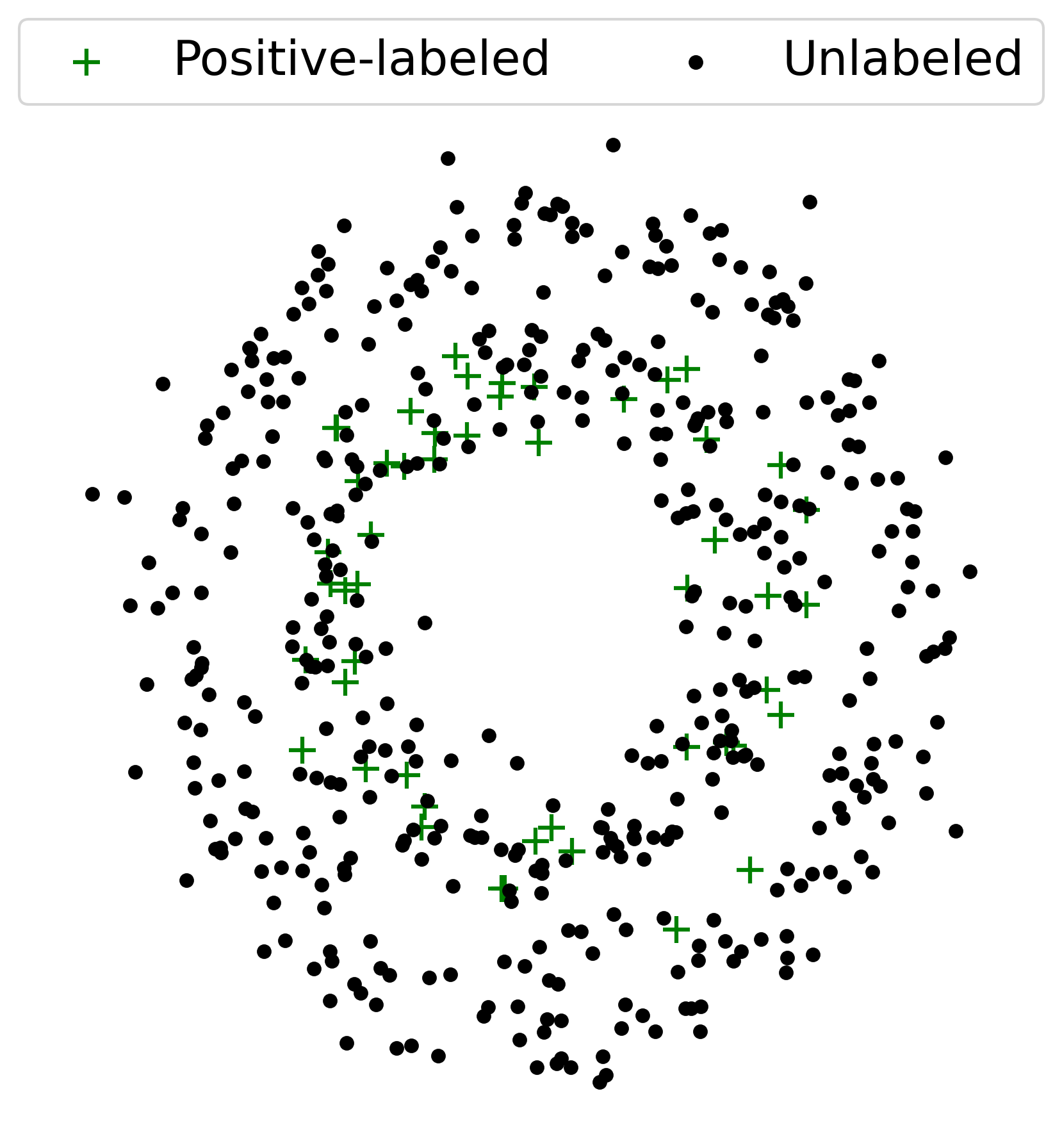}
\hfill
\caption{Toy dataset for illustrating the complete label knowledge (left) in typical supervised learning, and (right) in PU learning, where only some samples of a single label are given.}
\label{fig:pu_problem}
\end{figure}

Labeled data are often scarce and expensive to obtain in many real-world applications, making training machine-learning models a challenging task \cite{sevetlidis2022tackling}. In traditional supervised learning, the goal is to train a model to predict the correct class label for every sample in a training dataset \cite{abu2012learning}. The training data consist of labeled examples associated with a known class label. Typically, the class distribution of labeled data is assumed to be representative of the class distribution of unlabeled ones. Prior knowledge of the labels makes it easy to train a model to accurately predict the class labels for unseen samples. As \figurename~\ref{fig:pu_problem} shows, in the case of PU learning, the class label is known only for data belonging to a single class; thus, for negative samples, the label is unknown \cite{jaskie2022positive}. The lack of this knowledge makes it impossible to effectively train a typical binary classification model to distinguish between positive and negative classes. 

The unknown distribution of the negative samples renders the PU learning problem challenging \cite{hsieh2019classification}. This has led to the development of many different approaches, as described in the following section. The proposed novel methodology is boundary-aware and utilizes Gaussian sampling and anomaly detection. It creates a mass of embeddings from pairs of encoded positive-labeled data, which is essential for defining a rule-based boundary around the positive class. Negative samples are obtained from unlabeled data using anomaly detection. With the available positive-labeled data and the newly acquired negative set, the PU learning problem is simplified into a binary classification problem. A deep-learning binary classifier is employed to address this problem. Two datasets, CIFAR-10 \cite{krizhevsky2009learning} and Fashion-MNIST \cite{xiao2017fashion}, were used as the evaluation benchmarks. The proposed method achieved state-of-the-art results by following the same protocols as in \cite{zhao2022dist}.

In the following sections, related work in the field of PU learning is reviewed, and details of the steps involved in Dens-PU and experiments are described. The findings and implications of the results are discussed, followed by the potential impact of this approach. The paper concludes by demonstrating the effectiveness and usefulness of a boundary-aware PU learning methodology for image classification.


\section{Related work}


PU learning aims to learn a classifier $f$ that can accurately distinguish between the positive and negative classes despite the lack of labeled negative examples \cite{jaskie2022positive}. A generic approach to PU learning is to estimate the probability of each example in the set of unlabeled data $U$ belonging to the positive class and then use this probability estimate to train a binary classifier $f$ \cite{liu2002partially}.

Traditional binary classification algorithms can be adapted to handle the absence of negative-labeled data in PU learning, as discussed in \cite{li2003learning}. An unbiased risk estimator for PU learning, called uPU, was introduced in \cite{du2014analysis}, where the risk estimator may result negative values, which can be problematic owing to strong model overfitting. To address this, a non-negative risk estimator known as nnPU was proposed in \cite{kiryo2017positive}. Recently, a self-supervised extension of nnPU was introduced in \cite{chen2020self}, which utilizes auxiliary tasks, such as model calibration and a self-paced curriculum. Another variant of nnPU, presented in \cite{su2021positive}, modifies the weights to compensate for imbalanced data in the minority class. PUSB \cite{kato2019learning} relaxes the assumption of the order-preserving property, whereas, aPU \cite{hammoudeh2020learning} addresses the PU learning problem by fixing the negative distribution, while a positive distribution can shift arbitrarily.

PU learning approaches often rely on heuristic or statistical methods to estimate the negative class from unlabeled data, because identification of the negative class is challenging \cite{fung2005text,fusilier2015detecting}. Two-step methods, such as graph-based methods \cite{carnevali2021graph, wu2016positive}, differ in their approach for assigning labels to unlabeled data. PUbN assumes that the unlabeled data include a small number of negative examples that are highly representative of the negative class and combines them with positive-labeled data to train the model \cite{hsieh2019classification}. GenPU \cite{hou2017generative} leverages the GAN framework, whereas KLDCE \cite{gong2019loss} translates PU learning into a label noise problem and weakens its side effects via the centroid estimation of the corrupted negative set. PULNS \cite{luo2021pulns} incorporates reinforcement learning to select the effective negative samples. However, these approaches are prone to errors, and may have limited effectiveness in specific scenarios \cite{li2010negative}.

The identification of the negative class from the unlabeled data can be formulated as a problem of identifying samples that are unlikely to have been generated by the normal process of the underlying data distribution. Several anomaly detection methods have been proposed in the literature, including the Local Outlier Factor \cite{daneshpazhouh2013semi}, also LOF, which measures the degree of the local density of each point relative to its neighbors to detect the outliers, One-Class Support Vector Machines \cite{li2010positive} learns a hypersphere in the feature space that encompasses the majority of the data points and identifies the outliers as the points outside this hypersphere and Isolation Forest \cite{liu2008isolation} which discovers anomalies by calculating the degree of separation based on the steps needed to isolate a sample from its group \cite{hariri2019extended}.

However, all these methods have limitations in the context of PU learning. Anomaly detection methods rely on the assumption that the negative class is less frequent and distinctively different from the positive class. In practical scenarios, the negative class may be very similar to the positive one, making it difficult to detect outliers. Additionally, anomaly detection methods are known to be sensitive to the choice of hyperparameters and may require careful tuning to achieve optimal performance.

This study shares a similar intuition with \cite{basile2018density}, which suggests that density-based augmentation of the positive-labeled class can help distinguish negative samples. However, their approach differs from the proposed in this paper in that they utilize a probabilistic generative model to characterize the density distribution of the positive class. Another study, Dist-PU \cite{zhao2022dist} shares similarities with our study as it also aims to enhance the separability of the positive-negative distributions through entropy minimization and proposes using interpolation-based methods such as mixup to mitigate confirmation bias.

The novelty of the proposed approach lies in that unlike those that rely on biased negative data or variance-penalizing techniques, it generates biased positive samples to learn a boundary that encloses them and then estimates the degree of unlabeled samples belonging to the negative class, based on their distance from the learnt boundary.


\section{Methodology}

\begin{figure*}[t!]
\centering
\includegraphics[width=\textwidth]{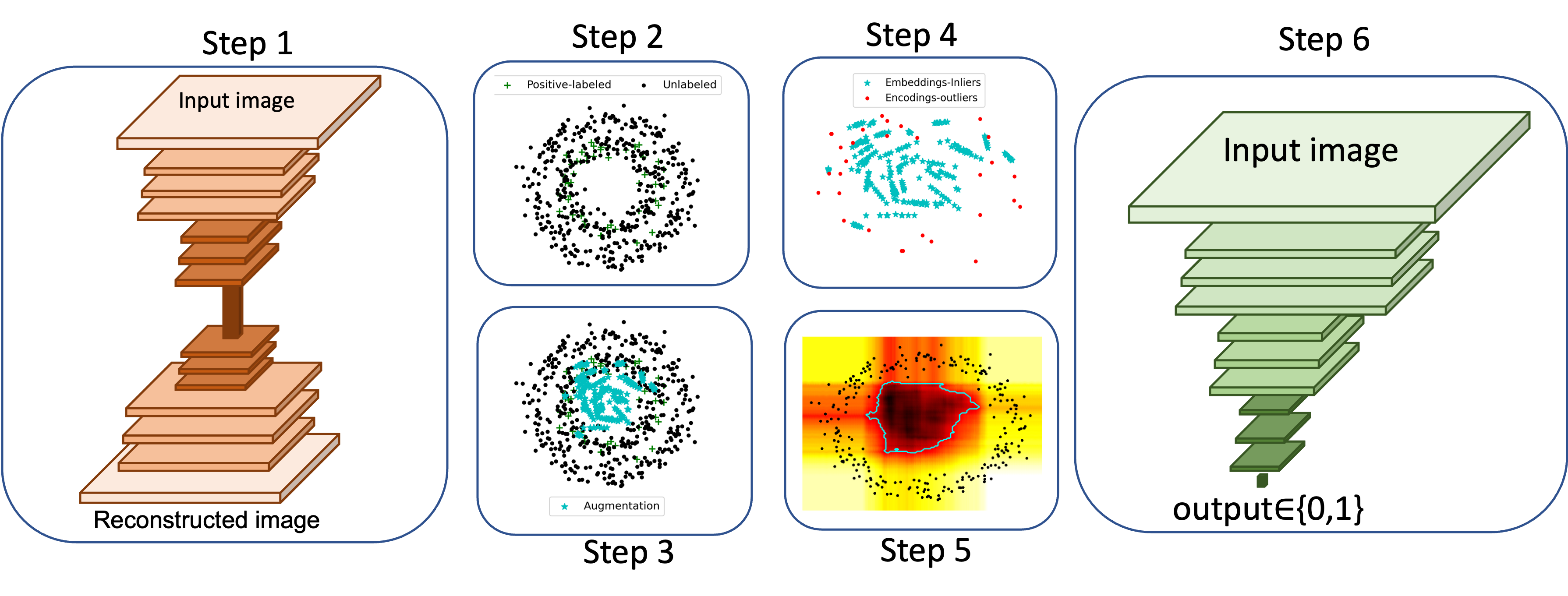}\hfill
\caption{The proposed methodology broken down in six steps.}
\label{fig:pipeline}
\end{figure*}

In a typical PU learning scenario, there is a set $X$ of samples $x_i\in X$ and $i=\{1,...,n\}$, a corresponding set $Y=\{0,1\}$ of labels $y_i \in Y$, and only one class is known for any given training subset of $X$. As a convention, $y=1$ is chosen as the known label such that $p(y=1|x)=1$ for some $x$, those considered the training set. Thus, there is no available information regarding the samples belonging to class $y=0$ during the training. In set notation, there are positive-labeled $P_L$ and $U$ unlabeled data, such that $X=P_L \cup U$. $U$ is composed of negative samples belonging to $N$ and the remaining unlabeled positive samples in $P_{UL}$, such that $U=P_{UL} \cup N$. PU learning aims to learn a classifier that correctly assigns both labels to unseen data using the available information.

The proposed methodology is motivated by the intuition that although we know nothing about the nature of $N$, knowing something about 
$P$ is sufficient for learning how to separate it from any other distribution. Although this is a perilous task, in this work, it is proposed that under the Selected Completely at Random Labeling Assumption (SCAR), it is possible to apply a rule-based boundary around the known distribution to separate it from dissimilar ones. Admittedly, this yields a highly biased model of $P$, but our aim is not to make $P_L$ larger with more positive-labeled samples; it is to draw a sample of $N$ from $U$ with confidence, such that the difficulty of the PU learning problem is reduced to a typical binary classification.

\subsection{High-level overview} 

A high-level overview of the proposed approach is shown in \figurename~\ref{fig:pipeline}. Specifically, Dens-PU exploits a Convolutional Autoencoder, which is trained on the positive-labeled data (step 1), to extract encodings using the latent space (step 2). The encodings are combined linearly in pairs to acquire new samples that lie between them (step 3). A dense mass is defined using the new samples as embeddings and the original encodings (step 4). This mass serves the purpose of delineating a boundary around the positive class, assuming that many data points outside the boundary are negative samples. This is where anomaly detection becomes relevant. Anomaly detection algorithms can identify data points that are significantly different from the majority of the data, allowing the separation of negative samples from the remaining unlabeled data (step 5). Thus, from the moment a negative class sample is obtained, the problem can be treated as any typical binary class classification in a supervised setting (step 6). The Dens-PU involves several techniques explained in the following sections.

\subsection{Convolutional Autoencoder}

A Convolutional Autoencoder (CAE) is a neural network architecture that is used for unsupervised learning. The network attempts to learn a transformed representation of the input data, in this case images, while reconstructing the original data as accurately as possible \cite{masci2011stacked}. A CAE consists of two parts: an encoder $z = f_{enc}(x) = \sigma(Wx + b)$ and a decoder $\hat{x} = f_{dec}(z) = \sigma(W'z + b')$, where $x$ is the input image, $W$,$W'$ are the weight matrices and $b$, $b'$ are the bias vectors of the encoder and decoder respectively, $\sigma(\cdot)$ is an activation function (such as ReLU), $z$ is the encoded representation, and $\hat{x}$ is the reconstructed image. 

The loss function used to train CAE is typically a measure of the difference between the input image and its reconstruction. The mean squared error (MSE) is a common choice for the reconstruction loss $L(x, \hat{x}) = \frac{1}{n}\sum_{i=1}^{n}(x_i - \hat{x}_i)^2$, where $n$ is the number of pixels in the image, $x_i$ is the $i$th pixel of the input image, and $\hat{x}_i$ is the corresponding pixel of the reconstructed image. A regularization term is frequently added to the loss function to prevent overfitting. One common choice is L\textsubscript{2} regularization and the total loss function becomes $L_{total} = L(x, \hat{x}) + L_{reg} = L(x, \hat{x}) + \frac{\lambda}{2}(\lVert W \rVert_2^2 + \lVert W' \rVert_2^2)$, where $L(x, \hat{x})$ is the reconstruction loss, and $L_{reg}$ is the regularization term.

\begin{figure}[t]
\centering
\includegraphics[width=0.33\textwidth]{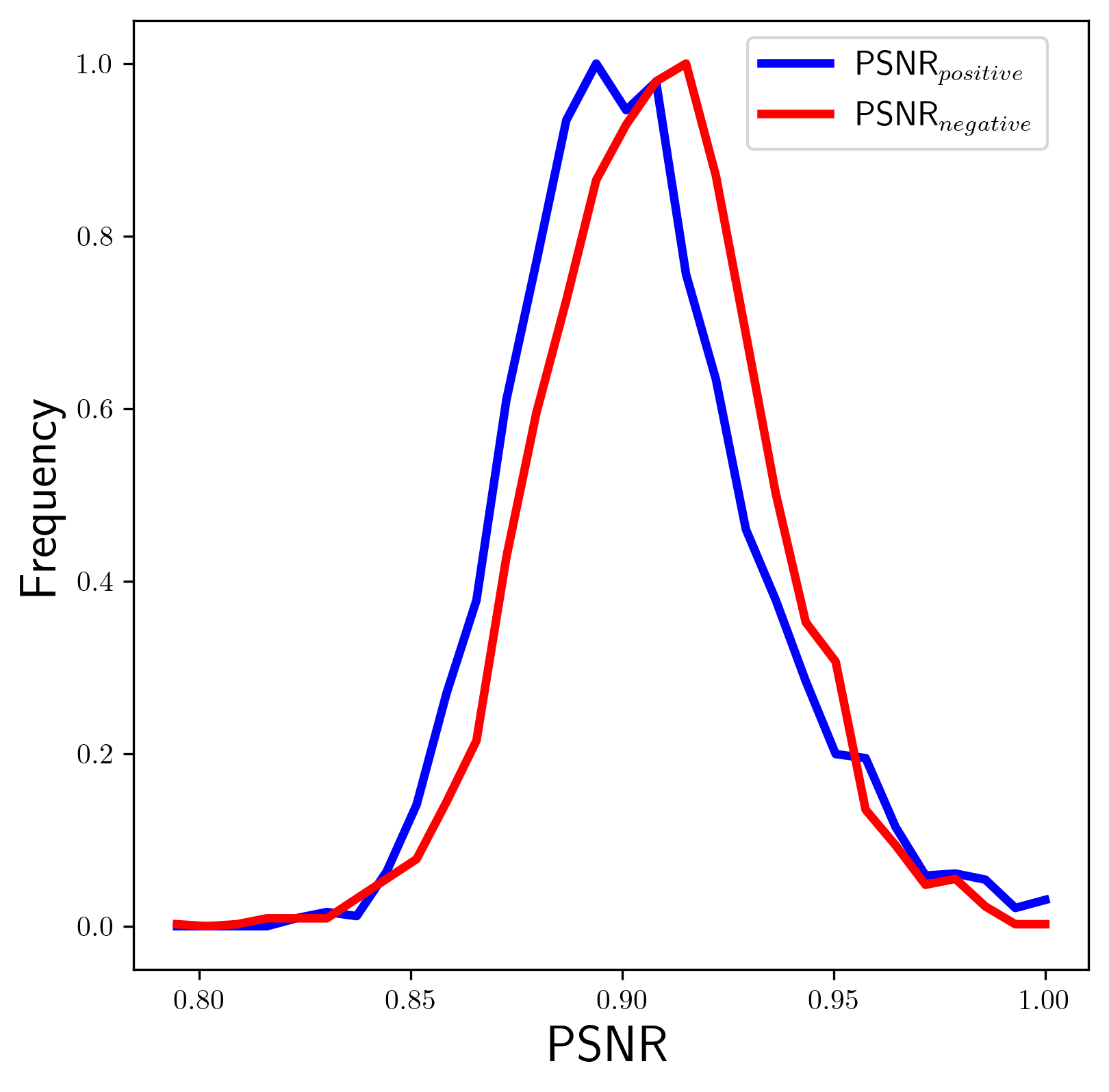}\hfill
\caption{The evaluation of the trained CAE using positive and negative data.}
\label{fig:psnr_distro_cae}
\end{figure}

CAE is the first block in the proposed methodology. It is trained on $P_L$ and is used to extract the encodings $Z_L$ and $Z_U$ from $P_L$ and $U$ respectively. Although one would expect a CAE to learn only class-specific representations, this is not true, since autoencoders are label agnostic and reconstruct any input globally. Nevertheless, any class-specific information is just an artifact retained within the weights as a result of thematically restricted input. In Dens-PU, the CAE learns how to reconstruct any natural image (despite labeling), given a sufficient number of input images. This is shown with a toy experiment in \figurename~\ref{fig:psnr_distro_cae} (left), where the distributions of PSNR values for positive (blue) and negative (red) samples from CIFAR-10\footnote{The experiment is as follows: a positive class was created by the union of \{\textit{automobile, airplane, ship, truck}\} classes, and a negative class by the union of \{\textit{bird, cat, deer, dog, frog, horse}\}. The CAE was trained using the positive class and was evaluated in reconstructing all images using the PSNR metric.} are shown. The observed heavy overlap is the reason that, even though these distributions are different\footnote{A Mann-Whitney statistical test with p-value$<$0.01 rejects the hypothesis of the two distributions being the same.}, they cannot be used for classification. Therefore, applying a simple threshold to the PSNR of two images --one with a known label and the other with an unknown-- to determine whether they belong to the same class is not a viable approach.

\subsection{Augmentation via Encoding Interpolation}

Data augmentation is an essential aspect of successful deep learning pipelines. This helps to mitigate sample size issues and often prevents overfitting. \textit{Mixup} is a recently introduced data augmentation technique that generates samples as random convex combinations of data points from a training set \cite{zhang2017mixup}. The label assignment of the newly created samples follows the distribution defined by the interpolation proportion. Studies have found that it significantly improves generalization in various tasks such as computer vision, natural language processing, and semi-supervised learning \cite{liang2018understanding}. 

Mixup modifies the Vicinal Risk Minimization \cite{chapelle2000vicinal}, in which the joint distribution $P(x,y)$ of a dataset $D=\{(x_i,y_i)\}^{n}_{i=1}$ is approximated by $P_\nu(\Tilde{x},\Tilde{y}) = \frac{1}{n}\sum^{n}_{i=1}\mu(\Tilde{x},\Tilde{y} | x_i, y_i)$, where $\mu$ is a \textit{vicinity distribution}. The vicinity distribution measures the probability of a \textit{virtual} feature pair $(\Tilde{x},\Tilde{y})$ being close to a \textit{training} pair. The \textit{mixup distribution} is introduced as follows:
\begin{equation} \label{eq:mixup}
    \mu(\Tilde{x},\Tilde{y} | x_i, y_i) = \frac{1}{n}\sum^{n}_{j}\mathop{\mathrm{E}}[\delta(\Tilde{x},\Tilde{y})]
\end{equation}
where $\Tilde{x} = \lambda \cdot x_i + (1- \lambda ) \cdot x_j, \Tilde{y} = \lambda \cdot y_i+(1- \lambda ) \cdot y_j
$, $\delta(x=x_i,y=y_i)$ is a Dirac mass centered at $(x_i,y_i)$ and $\lambda \sim Beta( \alpha , \alpha )$ for $\alpha \in (0,\infty)$. 

In the PU learning problem setting, the training data belong to the same class, which implies that the interpolated labels are also in the same class: $\Tilde{y}=\lambda \cdot y_i+(1- \lambda ) \cdot y_j=y_i$. Thus, \eqref{eq:mixup} becomes:
\begin{equation} \label{eq:densepu}
\mu(\Tilde{x} | x_i,y_i) = \frac{1}{n}\sum^{n}_{j}\mathop{\mathrm{E}}[\delta( \Tilde{x} = \lambda \cdot x_i + (1- \lambda ) \cdot x_j,y_i)]
\end{equation}
Therefore, the use of the distribution $P_\nu(\Tilde{x},\Tilde{y})$ around $\mu(\Tilde{x} | x_i,y_i)$, where $(x_i,1)\in P_L$, generates samples to augment the positive class. Note that $P_\nu$ contains the distribution $P_L$ because $\lambda \in [0, 1]$, especially for the cases where $\lambda$ indeed takes either $0$ or $1$. In this sense, the number of examples generated by sampling $P_\nu$ for any given pair $x_i$ and $x_j$, where $i \neq j$ and $x_i \neq x_j$, only makes the initial distribution denser.

\begin{figure}[t]
\centering
\includegraphics[width=0.33\textwidth]{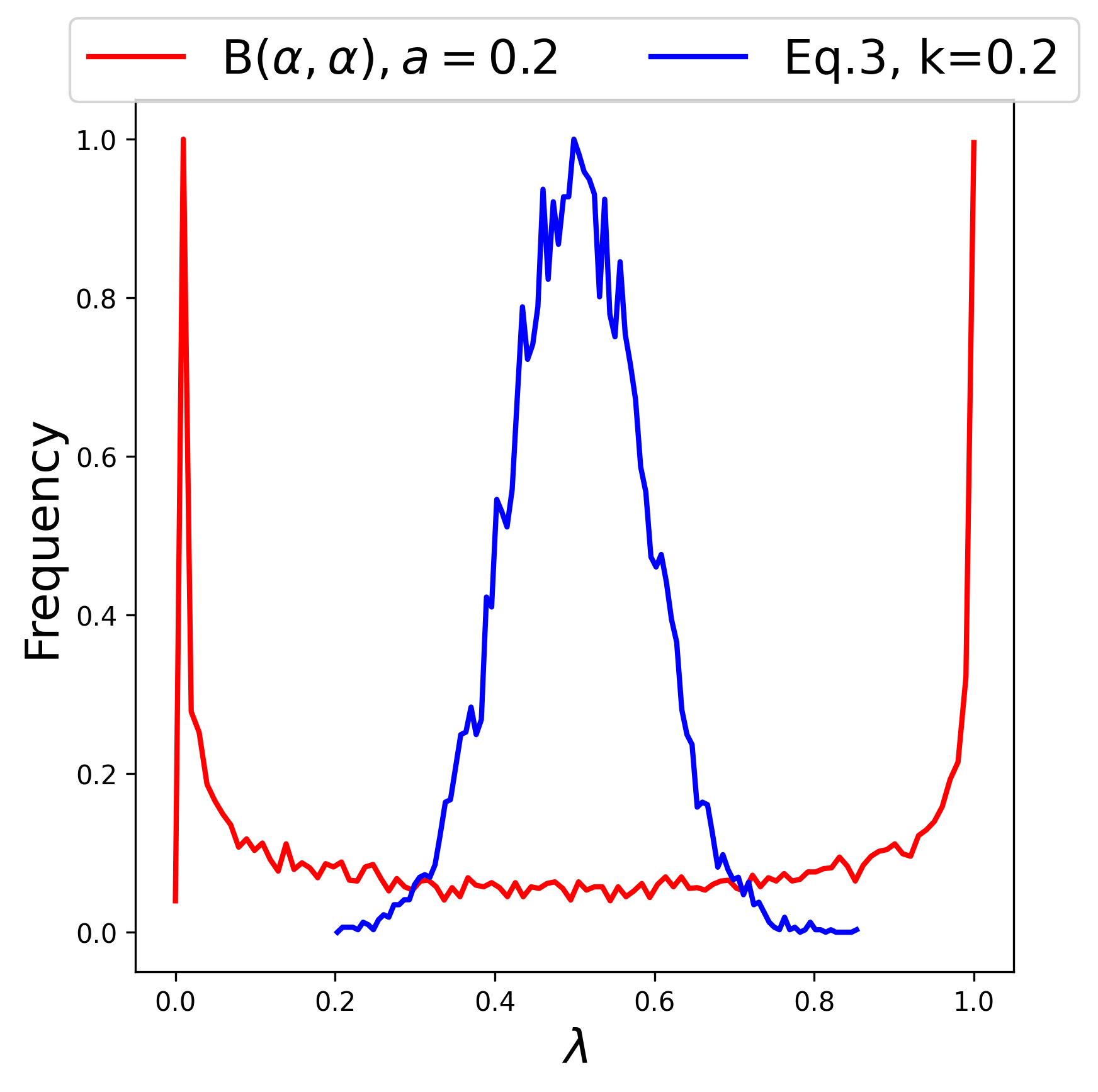}\hfill
\caption{The values $\lambda$ acquires for different distributions.}
\label{fig:lambda}
\end{figure}

In the context of the proposed Dens-PU, the set of encodings $Z_L$ received from the encoder of the CAE seed the creation of the set of embeddings $Z_\nu$, that follows a vicinity distribution. The proposed augmentation modifies the original mixup by (a) fixing the $y_i$ in \eqref{eq:densepu} owing to fact only one label known in PU-learning, (b) changing the sampling distribution for $\lambda$ from Beta into a Gaussian and (c) ensuring that the initial pair of samples are not considered in the produced samples ($\lambda \in (0,1)$). Specifically, $\lambda$ takes values from 

\begin{equation}\label{eq:lambda}
    \mathcal{N}=( \frac{z_j - z_i}{2} , e^k\cdot||\frac{z_j - z_i}{2}-z_i||_{2}^{2})
\end{equation}
where $z_i,z_j \in Z_L$ and the parameter $k\in(0,1)$ controls the spread of the variance, i.e., the similarity between the interpolated and the original samples. \figurename~\ref{fig:lambda} shows the different values $\lambda$ can take using the distribution from mixup (Eq.~\ref{eq:mixup} and the proposed \ref{eq:densepu}. Mixup focuses on staying close to the original samples whereas the proposed close to the center, avoiding the extreme values zero and one.


\begin{figure*}[!t]
\centering
\includegraphics[width=\textwidth]{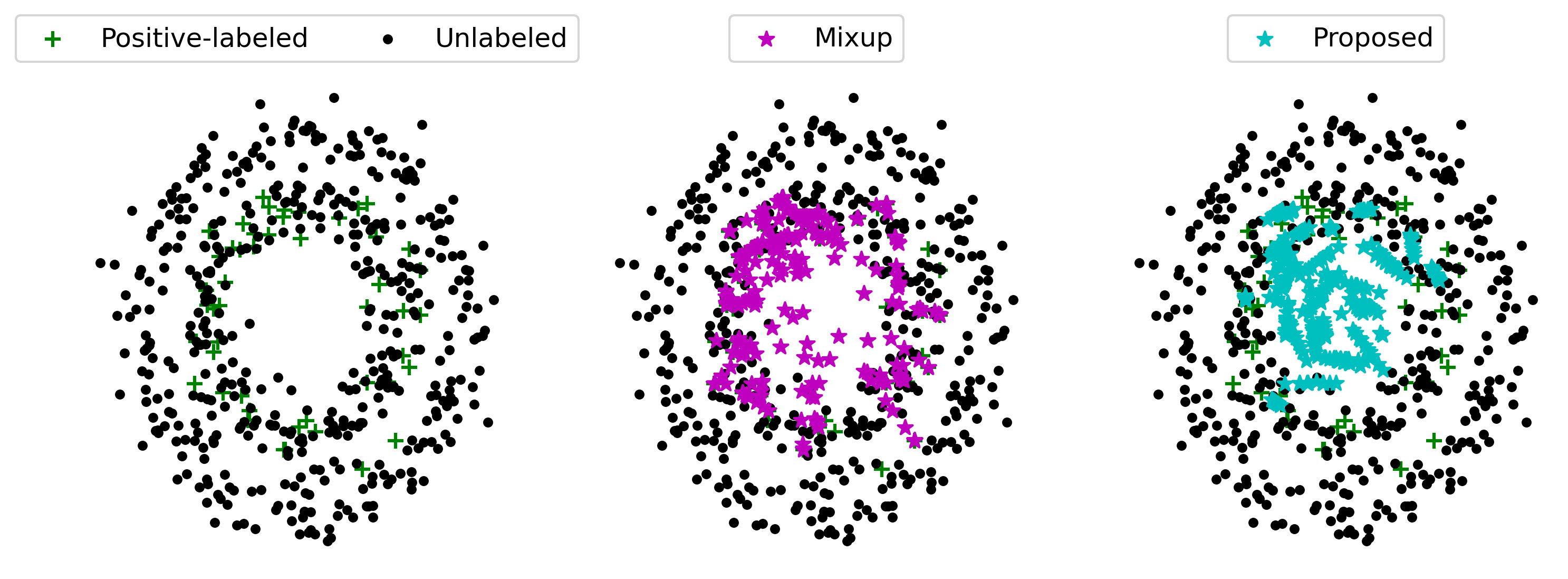}
\caption{A toy dataset (left) with positive (green) and unlabeled (black) samples, their mixup (middle) and the proposed augmentation (right).}
\label{fig:circles}
\end{figure*}

\figurename~\ref{fig:circles} shows a toy dataset consisting of positive (green) and unlabeled (black) samples (left), the result of mixup (middle) and the proposed density augmentation (right). The result of the augmentation are the magenta and cyan samples.


\subsection{Anomaly detection}

The problem in anomaly detection lies in identifying a subset of anomalous or outlier data points. Most methods rely on the assumption that the negative class is less frequent and distinctively different from the positive class. In other words, this can be formulated as a problem of identifying samples that are unlikely to have been generated by the normal process of the underlying data distribution.

Let $p$ be the probability density function of the embeddings i.e.,, the inliers and $q$ be the probability density function of the encodings i.e.,, the outliers. In addition, let $C$ be the contamination fraction, which is the proportion of the dataset expected to be outliers. The expected loss for anomaly detection is 
\begin{multline}\label{xxxxx}
L_E(\psi_{pred},\psi_{true}) = C \mathrm{E}{\chi\sim p(\chi)}[\psi_{pred}=1|\psi_{true}=0] +\\
+ (1-C) \mathrm{E}{\chi\sim q(\chi)}[\psi_{pred}=0|\psi_{true}=1]
\end{multline}
where the variable $\psi_{pred}$ represents the predicted label and $\psi_{true}$ represents the true label of sample $\chi$. The first term in the expected loss is the probability of a sample being misclassified as an inlier, which is equivalent to the probability of a false negative. The second term represents the probability that a point sampled from the inliers is misclassified as an outlier, which is equivalent to the probability of a false positive in the anomaly detection task.

At this stage the encodings $Z_L$ are labeled as outliers and the embeddings $Z_{\nu}$ inliers, owing to the significantly larger population of the embeddings. Hence, the contamination fraction $C=\frac{q}{p}=\frac{|Z_L|}{|Z_\nu|}$ can be estimated. This knowledge does not directly affect the form of the loss function, but helps to tune the $C$ constant. For example, if the fraction is small (i.e.,, the contamination rate is low), then the loss function may place more emphasis on minimizing false negatives (i.e.,, maximizing the true positive rate), because false positives are relatively rare. 


Overall, an anomaly detection method is used to learn the separation of $Z_\nu$ from $Z_L$ optimally. In practice, this is not a trivial task because $Z_\nu$ is an approximation of $Z_L$. Thus, parameter $C$ is expected to be near $0$ and the boundary between the two distributions to be very thin.

\subsection{From PU learning to Binary Classification}

Thus far, this methodology has transformed an input $x$ from $P_L$ into an encoding $z \in Z_L$ based on the latent representation of a CAE trained solely on data that lie in the same class. A simple yet novel augmentation method, which works similarly to the mixup, increased the density of the encodings, producing the embeddings. Labeling the embeddings as the inliers and the encodings as the outliers, allowed to define a boundary of the inlier class in the latent space using an anomaly detection method. 

Note that $U$ contains unlabeled samples belonging to either $P_{UL}$ or $N$; therefore, the corresponding encodings $Z_U=Z_{P_{UL}} \cup Z_N$ are obtained by the CAE. Unfortunately, the encodings do not provide sufficient information to separate $P_{UL}$ from $N$, as shown in \figurename~\ref{fig:psnr_distro_cae}. According to the framework of PU learning, only samples belonging to the positive class are labeled. Thus, the samples predicted as inliers were labeled $\Tilde{Z}_{PUL}$ and were subsequently removed from the unlabeled population, thus resulting in learning a data-driven boundary, using $Z_U - \Tilde{Z}_{P_{UL}}$. Although removing the predicted positive samples from the unlabeled ones might seem to imply that $\Tilde{Z}_{N} = Z_U - \Tilde{Z}_{P_{UL}}$ this is not explicitly true; technically, the leftovers in $Z_U - \Tilde{Z}_{P_{UL}}$ are still unlabeled samples. However, some information remains to be exploited, namely the degree of every leftover sample being an anomaly. Leftover samples reside outside of the learnt boundary, yet some samples are further than others. In this study $L_A$ is a sample's degree of being an anomaly, and is expressed as a rank defined by a sample's position within a sorted list of distances from the outermost samples in $Z_P\cup \Tilde{Z}_{P_{UL}}$, such that:
\begin{equation} \label{eq:anomaly_score}
 L_A(z_i)=\min_{\forall z_i\in Z_U - \Tilde{Z}_{P_{UL}}} d(z_i-z_j)    
\end{equation}
where $d(\cdot)$ is a metric function, $z_i \in Z_U - \Tilde{Z}_{P_{UL}}$, $z_j \in Z_P\cup \Tilde{Z}_{P_{UL}}$, and $l_n<l_{n+1}\forall l \in L_A$ for $n=\{1,2,..., |Z_U - \Tilde{Z}_{P_{UL}}|\}$. 
Any $\Tilde{N}$ subset of the ranked list \footnote{Practically, $\Tilde{N}$ can be as many as the initial $P_L$ samples.} can be used with confidence as a set of counter-examples to positive-labeled data. At the end of this process, $P_L$ and $\Tilde{N}$ are sets that can confidently be used as opposite datasets in a binary classification setting.

\begin{table*}[th!]
\centering
\setlength\tabcolsep{3pt}
\caption{Positive and Unlabeled dataset splits.}\label{table:datasets}
\begin{tabular}{ccccccc}
\hline
Dataset & Image Size & $|P_{L}|$ &  $|U|$ &  Testing & Positive Class              & Negative Class                      \\ \hline\hline
F-MNIST          & 28 x 28             & 1,000               & 59,000          & 10,000              & \{0, 2, 4, 6\}    & \{1, 3, 5, 7, 8, 9\} \\
CIFAR-10         & 32 x 32 x 3         & 1,000               & 49,000          & 10,000              & \{0, 1, 8, 9\} & \{2, 3, 4, 5, 6, 7\}           \\ \hline
\end{tabular}
\end{table*}

\begin{table*}[th!]
\centering
\setlength\tabcolsep{2pt}
\caption{Comparison of performances between the proposed method "Dens-PU" and the state of the art.}\label{table:results}
\begin{tabular}{c|c|cccccccccccccc|}
\hline
Dataset                   & Method           & \multicolumn{2}{c}{Acc (std)} &  & \multicolumn{2}{c}{Prec (std)} &  & \multicolumn{2}{c}{Rec (std)} &  & \multicolumn{2}{c}{F1 (std)} &  & \multicolumn{2}{c|}{AUC (std)} \\ \hline\hline
\multirow{11}{*}{\begin{turn}{90}F-MNIST\end{turn}} 
                          & uPU \cite{du2014analysis}
              & 94.20              & (0.3)     &  & 92.50               & (1.26)    &  & 92.59             & (0.8)     &  & 92.53             & (0.31)   &  & 97.34          & (0.54)        \\
                          & nnPU \cite{kiryo2017positive}
            & 94.44             & (0.49)    &  & 91.69              & (1.13)    &  & 94.69             & (0.84)    &  & 93.16             & (0.57)   &  & 97.53          & (0.48)        \\
                          & RP \cite{northcutt2017learning}
              & 92.37             & (1.08)    &  & 88.58              & (1.56)    &  & 92.94             & (2.38)    &  & 90.60              & (1.39)   &  & 97.14          & (0.58)        \\
                          & PUSB \cite{kato2019learning}
            & 94.5              & (0.36)    &  & 93.12              & (0.44)    &  & 93.12             & (0.44)    &  & 93.12             & (0.44)   &  & 97.31          & (0.5)         \\
                          & PUbN \cite{hsieh2019classification}
            & 94.82             & (0.16)    &  & 92.92              & (0.5)     &  & 94.24             & (0.93)    &  & 93.57             & (0.24)   &  & 94.72          & (0.29)        \\
                          & self-PU  \cite{chen2020self}
        & 94.75             & (0.25)    &  & 91.73              & (0.8)     &  & 95.50              & (0.61)    &  & 93.57             & (0.28)   &  & 97.62          & (0.31)        \\
                          & aPU  \cite{hammoudeh2020learning}
            & 94.71             & (0.34)    &  & 92.71              & (0.5)     &  & 94.20              & (1.06)    &  & 93.44             & (0.45)   &  & 97.67          & (0.4)         \\
                          & VPU \cite{chen2020variational}
             & 92.26             & (1.11)    &  & 89.04              & (2)       &  & 92.01             & (2)       &  & 90.48             & (1.35)   &  & 97.38          & (0.44)        \\
                          & ImbPU \cite{su2021positive}
           & 94.54             & (0.42)    &  & 92.81              & (1.53)    &  & 93.66             & (1.67)    &  & 93.21             & (0.52)   &  & 97.67          & (0.81)        \\
                          & Dist-PU \cite{zhao2022dist}         & 95.4              & (0.34)    &  & 94.18              & (0.9)     &  & 94.34             & (1)       &  & 94.25             & (0.43)   &  & 98.57          & (0.24)        \\ \cline{2-16} 
                          & \textbf{Dens-PU} & \textbf{95.73}    & (0.53)       &  & 92.36              & (1.1)       &  & \textbf{94.8}    & (0.6)       &  & \textbf{94.80}     & (0.62)      &  & 96.00             & (0.73)           \\ \hline\hline
\multirow{11}{*}{\begin{turn}{90}CIFAR-10\end{turn}} 
                          & uPU  \cite{du2014analysis}
            & 88.35             & (0.45)    &  & 87.18              & (2.39)    &  & 83.23             & (2.68)    &  & 85.10              & (0.31)   &  & 94.91          & (0.62)        \\
                          & nnPU \cite{kiryo2017positive}
            & 88.89             & (0.45)    &  & 86.18              & (1.15)    &  & 86.05             & (1.42)    &  & 86.10              & (0.57)   &  & 95.12          & (0.52)        \\
                          & RP  \cite{northcutt2017learning}
             & 88.73             & (0.15)    &  & 86.01              & (1.01)    &  & 85.82             & (1.51)    &  & 85.9              & (1.39)   &  & 95.17          & (0.23)        \\
                          & PUSB \cite{kato2019learning}
            & 88.95             & (0.41)    &  & 86.19              & (0.51)    &  & 86.19             & (0.5)     &  & 86.19             & (0.44)   &  & 95.13          & (0.52)        \\
                          & PUbN  \cite{hsieh2019classification}
           & 89.83             & (0.3)     &  & 87.85              & (0.98)    &  & 86.56             & (1.87)    &  & 87.18             & (0.24)   &  & 89.28          & (0.54)        \\
                          & self-PU  \cite{chen2020self}
        & 89.28             & (0.72)    &  & 86.16              & (0.78)    &  & 87.21             & (2.35)    &  & 86.67             & (0.28)   &  & 95.47          & (0.58)        \\
                          & aPU  \cite{hammoudeh2020learning}
            & 89.05             & (0.52)    &  & 86.29              & (1.3)     &  & 86.37             & (0.79)    &  & 86.32             & (0.45)   &  & 95.09          & (0.42)        \\
                          & VPU  \cite{chen2020variational}
            & 87.99             & (0.58)    &  & 86.72              & (1.41)    &  & 82.71             & (2.84)    &  & 84.63             & (1.35)   &  & 94.51          & (0.41)        \\
                          & ImbPU \cite{su2021positive}
           & 89.41             & (0.46)    &  & 86.69              & (0.9)     &  & 86.87             & (0.82)    &  & 86.77             & (0.52)   &  & 95.52          & (0.27)        \\
                          & Dist-PU \cite{zhao2022dist}         & 91.88             & (0.52)    &  & 89.87              & (1.09)    &  & 89.84             & (0.8)     &  & 89.85             & (0.43)   &  & 96.92          & (0.45)        \\ \cline{2-16} 
                          & \textbf{Dens-PU} & \textbf{93.63}    & (0.44)       &  & \textbf{92.68}     & (1.31)       &  & \textbf{91.25}    & (1.12)       &  & \textbf{91.96}    & (0.8)      &  & 93.22         & (0.99)           \\ \hline
\end{tabular}
\end{table*}

\section{Experiments}

\subsection{Configuration}

In the experiments conducted\footnote{Dens-PU was implemented in Python with Keras as the deep-learning backend.} for the validation of the proposed methodology, two well-known datasets have been used, namely CIFAR-10 \cite{krizhevsky2009learning} and Fashion-MNIST \cite{xiao2017fashion}. For the purposes of the PU learning setting, transportation related images were merged into one class and living things related images into the second, for the case of CIFAR-10, whereas top clothes were merged into one class and the rest images in the other one, for the case of Fashion-MNIST (see \figurename~\ref{table:datasets}). \tablename~\ref{table:datasets} shows how the datasets were handled.

To keep the hyperparameters of the algorithms and deep learning architectures the same, the images from Fashion-MNIST were upscaled to match the 32 × 32 size of CIFAR-10. Moreover, two additional channels were added such that a grayscale image from the Fashion-MNIST be treated as an RGB image, by replicating the single channel. 

The CAE was a symmetrical autoencoder with three convolutional and three deconvolutional layers. The architecture was kept as simple as possible, because obtaining only latent encodings was desirable. The filters for each encoding layer were 64, 32, and 8, respectively, and the reverse order was applied to the decoder. MaxPooling halved the shape of the filters between each convolution. The encoding extracted from the latent space has dimensions $Z\in \mathbb{R}^{1\times512}$. The training batch size was set to 64 for all datasets. Adam \cite{kingma2014adam} was chosen as the optimizer. MSE acted as a loss function. The initial learning rate and weight decay were set as $10^{-4} $ and $10^{-3} $, respectively. The architecture is trained for $50$ epochs.

The proposed interpolation \eqref{eq:lambda} had three parameters to be configured: (a) the proximity towards the midpoint controlled by $k=0.2$, (b) the sample size \footnote{from the possible pairs $\binom{|Z_L|}{2}$, which corresponds to 99\% confidence with a 1\% margin of error} of positive pairs $|(z_i,z_j)|=16,000$ which were sampled with no replacement and were used as the extreme points of the line segment made by said interpolation, and (c) the number of interpolated samples that belong to the line segment of a given pair was $s=11$.

As an anomaly detection method, Isolation Forest was chosen because it is fast to train owing to its random splits, adapts well in highly non-linear spaces, and it is easy to optimize its hyper-parameters \cite{liu2008isolation, hariri2019extended, staerman2019functional}. Hence, the number of estimators was set to $1000$, the depth was left at the default operation, the number of samples per estimator was $256$ and the contamination fraction at $C=\frac{|Z_L|}{s\cdot|Z_\nu|}\approx 0.005$. Conveniently, Isolation Forest uses the number of steps needed to separate a sample as an anomaly score. Dens-PU uses this score to replace $d$ in \eqref{eq:anomaly_score} and to calculate the ordered list $L_A$.

Finally, the VGG-16 architecture was chosen as a binary classifier for all the experiments. This decision was made because VGG-16 is not a complex deep classifier, yet it is capable of performing well in typical classification tasks. The weights were randomly initialized. Average pooling and two dense layers are added after the last VGG block. The dense layers consisted of 128 neurons with ReLU activation, connected to a single neuron and sigmoid activation. The optimizer was SGD with a learning rate and weight decay of $10^{-4} $ and $10^{-3}$ respectively. The model was trained for 200 epochs or until it reached a plateau with a batch size of 32 samples.

\begin{table*}[t!]
\centering
\setlength\tabcolsep{1pt}
\caption{Assessing $F_1$-score performance for varying initial populations of known samples in $P_L$.} \label{table:ablation_percentage}
\begin{tabular}{lllllll}
\hline
                              & 1\% & 5\% & 10\% & 25\% & 30\% & 50\% \\ \hline\hline
F-MNIST  &   88.71 (±0.67)  & 93.83 (±0.74)   &  94.35 (± 0.93)    &   94.71 (±0.80)   &   94.68 (±0.68)   &  95.1 (±0.87)     \\
CIFAR-10 & 82.85 (±1.8)    & 89.65 (±0.90)     &  92.23 (±0.56)     & 92.02 (±0.72)      &  92.55 (±1.03)     &  93.79 (±0.96)     \\ \hline
\end{tabular}
\end{table*}

\subsection{Results}

\tablename~\ref{table:results} presents the evaluation results for ten PU-learning methods on two datasets, Fashion-MNIST and CIFAR-10. The $F_1$-score, Area Under the Curve (AUC), Accuracy, Precision, and Recall metrics \cite{jaskie2022positive} were used to evaluate the performance of each algorithm for both datasets. 

For Fashion-MNIST, Dens-PU outperformed all traditional methods in terms of F1-score, Precision, Recall, and accuracy metrics. Among the state-of-the-art methods, Dist-PU and ImbPU exhibited the best performance. For CIFAR10, Dens-PU also outperformed all the traditional methods in the same two metrics. The performance of the state-of-the-art methods was comparable, with Dist-PU, nnPU, and ImbPU attaining the highest scores. Self-PU and PUbN were among the most competitive baselines because of their extra designs, such as mentor nets and pre-trained models. The VPU method, which does not use class prior information, shows relatively less promising results than other baselines.

\subsection{Ablation studies}


Before moving on to the presentation of the ablations studies, it is important to note that minibatch training was used in all cases of unbalanced class ablation study. A subset of samples from the larger class is selected to match the size of the smaller class. Failure to do so would result in the classifier heavily overfitting the larger class, leading to an extremely low $F_1$-score. However, the relatively high performance observed may be attributed to learning with noisy labels \cite{ghosh2017robust}, which is beyond the scope of this work. Instead of using minibatch training when having uneven classes, other approaches involve learning a weighted classifier \cite{zhou2012multi} or applying a weighted loss\cite{ren2018learning}, but these are not explored here as well.

The ablation begins with the investigation of the impact of different sizes of positive-labeled sample populations at the beginning of the method, denoted as $|P_L|$. The $F_1$-score is used to measure the performance, and \tablename~\ref{table:ablation_percentage} shows the results for both the CIFAR-10 and Fashion-MNIST datasets. The findings reveal that the performance of the proposed methodology remains unaffected for an initial sample size of $5-30\%$ of the training dataset. When limiting the initial information to only $1\%$, the performance drops significantly below that of the state-of-the-art, although classification remains feasible within acceptable accuracy limits ($F_1$-score remains high). Conversely, when half of the training dataset is available, the performance improves; however, such large amounts are not usually available in typical PU learning scenarios.

A second ablation study examined the effect of the two parameters controlling (a) the data used for learning the data-driven boundary and (b) the criterion regarding counter-example selection. Regarding (a) three modes for the selection of $Z_\nu$ were considered, (i) using only the available encodings of positive-labeled samples with no other information, (ii) using the encodings of positive-labeled samples and embeddings from mixup, and (iii) using the encodings of positive-labeled samples and embeddings generated by the proposed interpolation. Regarding (b) random sampling was selected, denoted by $\in_R$. The reason for evaluating Dens-PU on these particular parameters is that this work proposes to be performed differently than others, and they are considered the main contributions. \tablename~\ref{table:ablation_variants} shows the results of this test, where $F_1$-score was used as the evaluation metric. Variant $1$ demonstrates a naive binary classification approach in which the unlabeled set is used directly as a counter-example to the positive-labeled. Variants $1$ to $3$ exhibit the worst performance compared with the other variants, which is expected because the data-driven boundary no longer reflects its intended design purposes. A density augmentation technique is required to construct a boundary around the approximated positive-labeled class. Variants $6$ and $9$ demonstrate that using an anomaly score as proposed is a logical criterion for selecting samples to be assigned to the counter-examples class $\Tilde{N}$ compared with random sampling from the unlabeled set (i.e.,, variants 4 and 6) and random sampling from the leftovers (i.e.,, variants 5 and 7). As \eqref{eq:mixup} (mixup) and \eqref{eq:densepu} (proposed) are similar, variants $4,5$ and $6,7$ show comparable performance, with the second group exhibiting more stable accuracy. Overall, variant $7$ outperformed all others, indicating that the proposed modules provide an additional advantage over the existing.

\begin{table}[t!]
\centering
\setlength\tabcolsep{5pt}
\caption{Evaluating the $F_1$-score performance of the proposed Dens-PU and its variants on CIFAR-10.} \label{table:ablation_variants}
\begin{tabular}{cccc}
\hline
  Variant & $Z_{\nu}$ & Selection of $\Tilde{N}$  & $F_1$  \\\hline\hline
  1 & -      & $x\in_RU$  &   80.46 (±0.18) \\
  2 & ${Z_P}$      & $x\in_R U- \Tilde{P}_{PUL}$ & 86.74   (±0.41)  \\
  3 & ${Z_P}$      & \eqref{eq:anomaly_score} &   83.02 (±1.89) \\
  4 & MixUp \cite{zhang2017mixup}     & $x\in_R U- \Tilde{P}_{PUL}$   &  90.38 (±0.71) \\
  5 & MixUp \cite{zhang2017mixup}    & \eqref{eq:anomaly_score}  &  90.51 (±0.63) \\
  6 & \eqref{eq:densepu}  & $x\in_R U- \Tilde{P}_{PUL}$ &  89.88 (±1.10)  \\ \hline    
  7 & \eqref{eq:densepu}  & \eqref{eq:anomaly_score} &  \textbf{91.20 (±0.97)} \\ \hline 
\end{tabular}
\end{table}

A final ablation study considered the different population sizes of the counter-example set $\Tilde{N}$. Three variants were examined, with the counter-example set size being (1) equal to the leftovers (i.e.,, $|\Tilde{N}| = |U-P_{PUL}|$), effectively choosing all remaining samples, (2) a random number within the leftover size limits, and (3) equal to the size of the positive-labeled set (i.e.,, $|\Tilde{N}| = |P_L|$). The performance was evaluated using the $F_1$-score in \tablename~\ref{table:ablation_population}. As expected, the first variant performs the worst compared to the other two because selecting all leftovers means mistakenly acquiring positive samples predicted as anomalies. Moreover, the leftover size can be larger than the positive-labeled set (i.e.,, $|U-P_{PUL}|\gg|P_L|$), causing imbalanced classes and resulting in a poor binary classifier performance. Selecting a random number exhibits similar risks; however, this time, the counter-example set can be smaller than the positive-labeled set. The third variant confirms that balancing the classes is the best option.

\begin{table}[t!]
\centering
\setlength\tabcolsep{4pt}
\caption{Evaluating the F1-score performance for different counter-example populations on F-MNIST (F) and CIFAR-10(C).} \label{table:ablation_population}
\begin{tabular}{cccc}
\hline
Variant & $|\Tilde{N}|$     & $F_1^{(F)}$ & $F_1^{(C)}$\\ \hline\hline
1       & $|U-P_{PUL}|$      &  91.73 (±0.31)  & 89.80 (±0.84)\\
2       & $ \in_R |U-P_{PUL}|$  &  88.91 (±3.01)  & 84.12 (±1.13)\\
3       & $|P_L|$            &  \textbf{93.41 (±0.62)}  & \textbf{91.33 (±0.75)}\\ \hline
\end{tabular}
\end{table}

\section{Discussion}

The experiments conducted in this study demonstrate that the proposed approach effectively addresses the problem of PU learning. The results confirmed that density augmentation plays a critical role in PU learning. Overall, the proposed approach outperforms ten reference methods in terms of accuracy, precision, recall, and $F_1$-score. Although several components were combined, Dens-PU was easy to set up and no special computational resources were required for the experiments.

One drawback of the proposed methodology is its dependence on the quality of encodings extracted from positive-labeled data. The performance may be affected if the encodings contain noisy or irrelevant features. In addition, the choice of Gaussian sampling parameter $k$ can affect the quality of the created samples. For example, if $k$ is set close to $0$, all samples may look like the midpoint, reducing them to a single sample. If $k$ is set too high, the boundary between the positive-labeled distribution and its approximation may not be defined, leading to lower performance in the PU learning task.

\section{Conclusion}

This study addressed the problem of PU learning by introducing a novel method, Dens-PU, that learns a data-driven boundary around the approximated distribution of positive-labeled data. It uses an anomaly detection algorithm to discover negative samples with confidence, and once obtained, a typical supervised learning binary classification can be performed.

Dens-PU was extensively tested against reference methods from the relevant literature, using benchmark image datasets, and the results showed a significant performance improvement achieving state-of-the-art accuracy with respect to the $F_1$-score. Potentially, it can be applied for classification scenarios that labeled data are expensive or difficult to obtain, like medical imaging, fraud detection and dataset creation pipelines.

Future studies can investigate other density augmentation methods or improve the classification results by exploring training strategies for unbalanced classes, for example, by applying weighted loss or learning a weighted classifier such that the population of the counter-example set can be increased. Dens-PU uses encodings to draw a negative sample from the unlabeled data, therefore applying it to other modalities like text and audio might be feasible.

{\small
\bibliographystyle{ieeetr}
\bibliography{paper}
}

\end{document}